\documentclass[journal]{IEEEtran}

\usepackage[mathscr]{eucal}
\usepackage[cmex10]{amsmath}
\usepackage{epsfig,epsf,psfrag}
\usepackage{amssymb,amsmath,amsthm,amsfonts,latexsym}
\usepackage{amsmath,graphicx,bm,xcolor,url}
\usepackage[caption=false]{subfig} 
\usepackage{fixltx2e}
\usepackage{array}
\usepackage{verbatim}
\usepackage{bm}
\usepackage{algorithmic, cite}
\usepackage{algorithm}
\usepackage{verbatim}
\usepackage{textcomp}
\usepackage{mathrsfs}
\usepackage{epstopdf}

\catcode`~=11 \def\UrlSpecials{\do\~{\kern -.15em\lower .7ex\hbox{~}\kern .04em}} \catcode`~=13 

\allowdisplaybreaks[3]

\newcommand{\calI}{\mathcal{I}}

\newcommand{\calL}{\mathcal{L}}

\newcommand{\calN}{\mathcal{N}}
\newcommand{\calO}{\mathcal{O}}

\newcommand{\calR}{\mathcal{R}}
\newcommand{\calS}{\mathcal{S}}

\newcommand{\ba}{\mathbf{a}}
\newcommand{\bA}{\mathbf{A}}
\newcommand{\bb}{\mathbf{b}}
\newcommand{\bB}{\mathbf{B}}
\newcommand{\bc}{\mathbf{c}}
\newcommand{\bC}{\mathbf{C}}
\newcommand{\bd}{\mathbf{d}}
\newcommand{\bD}{\mathbf{D}}
\newcommand{\be}{\mathbf{e}}

\newcommand{\bg}{\mathbf{g}}

\newcommand{\bh}{\mathbf{h}}
\newcommand{\bH}{\mathbf{H}}

\newcommand{\bI}{\mathbf{I}}

\newcommand{\bL}{\mathbf{L}}

\newcommand{\bM}{\mathbf{M}}

\newcommand{\bP}{\mathbf{P}}
\newcommand{\bq}{\mathbf{q}}
\newcommand{\bQ}{\mathbf{Q}}

\newcommand{\bR}{\mathbf{R}}
\newcommand{\bs}{\mathbf{s}}
\newcommand{\bS}{\mathbf{S}}

\newcommand{\bU}{\mathbf{U}}
\newcommand{\bv}{\mathbf{v}}
\newcommand{\bV}{\mathbf{V}}

\newcommand{\bz}{\mathbf{z}}

\DeclareMathAlphabet{\mathbsf}{OT1}{cmss}{bx}{n}
\DeclareMathAlphabet{\mathssf}{OT1}{cmss}{m}{sl}

\DeclareSymbolFont{bsfletters}{OT1}{cmss}{bx}{n}  
\DeclareSymbolFont{ssfletters}{OT1}{cmss}{m}{n}
\DeclareMathSymbol{\bsfGamma}{0}{bsfletters}{'000}
\DeclareMathSymbol{\ssfGamma}{0}{ssfletters}{'000}
\DeclareMathSymbol{\bsfDelta}{0}{bsfletters}{'001}
\DeclareMathSymbol{\ssfDelta}{0}{ssfletters}{'001}
\DeclareMathSymbol{\bsfTheta}{0}{bsfletters}{'002}
\DeclareMathSymbol{\ssfTheta}{0}{ssfletters}{'002}
\DeclareMathSymbol{\bsfLambda}{0}{bsfletters}{'003}
\DeclareMathSymbol{\ssfLambda}{0}{ssfletters}{'003}
\DeclareMathSymbol{\bsfXi}{0}{bsfletters}{'004}
\DeclareMathSymbol{\ssfXi}{0}{ssfletters}{'004}
\DeclareMathSymbol{\bsfPi}{0}{bsfletters}{'005}
\DeclareMathSymbol{\ssfPi}{0}{ssfletters}{'005}
\DeclareMathSymbol{\bsfSigma}{0}{bsfletters}{'006}
\DeclareMathSymbol{\ssfSigma}{0}{ssfletters}{'006}
\DeclareMathSymbol{\bsfUpsilon}{0}{bsfletters}{'007}
\DeclareMathSymbol{\ssfUpsilon}{0}{ssfletters}{'007}
\DeclareMathSymbol{\bsfPhi}{0}{bsfletters}{'010}
\DeclareMathSymbol{\ssfPhi}{0}{ssfletters}{'010}
\DeclareMathSymbol{\bsfPsi}{0}{bsfletters}{'011}
\DeclareMathSymbol{\ssfPsi}{0}{ssfletters}{'011}
\DeclareMathSymbol{\bsfOmega}{0}{bsfletters}{'012}
\DeclareMathSymbol{\ssfOmega}{0}{ssfletters}{'012}

\DeclareMathOperator{\sgn}{sgn}

\DeclareMathOperator{\rank}{rank}

\newtheorem{theorem}{Theorem} 
\newtheorem{lemma}[theorem]{Lemma}

\newtheorem{definition}{Definition}

\newtheorem{remark}{Remark}

\newtheorem{assumption}{Assumption}
\newtheorem{data model}{Data Model}
\newtheorem{analytical problem}{Analytical Problem}

\newcommand{\qednew}{\nobreak \ifvmode \relax \else
      \ifdim\lastskip<1.5em \hskip-\lastskip
      \hskip1.5em plus0em minus0.5em \fi \nobreak
      \vrule height0.75em width0.5em depth0.25em\fi}

\begin{document}
\title{Low Rank Matrix Recovery with Simultaneous Presence of Outliers and Sparse Corruption} 

\author{Mostafa~Rahmani, \IEEEmembership{Student Member,~IEEE} and George~K.~Atia,~\IEEEmembership{Member,~IEEE} 
\thanks{This work was supported in part by NSF CAREER Award CCF-1552497 and NSF Grant CCF-1320547.

The authors are with the Department of Electrical and Computer Engineering, University of Central Florida, Orlando, FL 32816 USA (e-mails: mostafa@knights.ucf.edu, george.atia@ucf.edu).}
}

\markboth{}%
{Shell \MakeLowercase{\textit{et al.}}: Bare Demo of IEEEtran.cls for Journals}
\maketitle

\begin{abstract}
We study a data model in which the data matrix $\bD \in \mathbb{R}^{N_1 \times N_2}$ can be expressed as
\begin{eqnarray}
\bD = \bL + \bS + \bC \:,
\label{eq:model}
\end{eqnarray}
where $\bL$ is a low rank matrix, $\bS$ an element-wise sparse matrix and $\bC$ a matrix whose non-zero columns are outlying data points.
To date, robust PCA algorithms have solely considered models with either $\bS$ or $\bC$, but not both. As such, existing algorithms cannot account for simultaneous element-wise and column-wise corruptions. In this paper,  a new robust PCA algorithm that is robust to simultaneous types of corruption is proposed. Our approach hinges on the sparse approximation of a sparsely corrupted column so that the sparse expansion of a column with respect to the other data points is used to distinguish a sparsely corrupted inlier column from an outlying data point. We also develop a randomized design which provides a scalable implementation of the proposed approach. The core idea of sparse approximation is analyzed analytically where we show that the underlying $\ell_1$-norm minimization can obtain the representation of an inlier in presence of sparse corruptions.

\end{abstract}


\IEEEpeerreviewmaketitle

\section{Introduction}
Standard tools such as Principal Component Analysis (PCA) has been routinely used to reduce dimensionality by finding linear projections of high-dimensional data into lower dimensional subspaces. The basic idea is to project the data along the directions where it is most spread out so that the residual information
loss is minimized. This has been the basis for much progress
in a broad range of data analysis problems, including problems in computer vision, communications, image processing, machine learning and bioinformatics \cite{lamport44,lamport45,cogan2017multi,hosseini2016real,hajinezhad2015nonconvex}.

PCA is notoriously sensitive to outliers, which prompted substantial effort in developing robust algorithms that are not unduly affected by outliers. Two distinct robust PCA problems were considered in prior work depending on the underlying data corruption model, namely, the low rank plus sparse matrix decomposition \cite{lamport1,lamport22} and the outlier detection problem \cite{lamport10}.

\subsection{Low rank plus sparse matrix decomposition}
In this problem, the data matrix $\bD$ is a superposition of a low rank matrix $\bL$ representing the low dimensional subspace, and a sparse component $\bS$ with arbitrary support, whose entries can have arbitrarily large magnitude modeling element-wise data corruption \cite{lamport22,lamport1,lamport29,zhou2010stable,minaeee2015screen,rahmani2016robust}, i.e., $$\bD = \bL + \bS.$$
%
For instance, \cite{lamport22} assumes a Bernoulli model for the support of $\bS$ in which each element of $\bS$ is non-zero with a certain small probability. Given the arbitrary support, all the columns/rows of $\bL$ may be affected by the outliers.
The cutting-edge Principal Component Pursuit (PCP) approach developed in \cite{lamport1} and \cite{lamport22} directly decomposes $\bD$ into its low rank and sparse components by solving a convex program that minimizes a weighted combination of the nuclear norm $\|\dot{\bL}\|$(sum of singular values), and the $\ell_1$-norm $\|\dot{\bS}\|_1$,
\begin{eqnarray}
\begin{aligned}
& \underset{\dot{\bL},\dot{\bS}}{\min}
& & \|\dot{\bL} \|_* + \lambda\|\dot{\bS}\|_1 \\
& \text{subject to}
& & \dot{\bL} + \dot{\bS} = \bD.
\end{aligned}
\label{eq: PCP}
\end{eqnarray}
If the column and row spaces of $\bL$ are sufficiently incoherent and the non-zero elements of $\bS$ sufficiently diffused, (\ref{eq: PCP}) can provably recover the exact low rank and sparse components \cite{lamport1,lamport22}.


\subsection{Outlier detection}
In the outlier detection problem, outliers only affect a portion of the columns of $\bL$, i.e., corruption is column-wise. The given data is modeled as $$\bD = \bL + \bC.$$
A set of the columns of the outlier matrix $\bC$, the so-called outliers, are non-zero and they do not lie in the Column Space (CS) of $\bL$. In this problem, it is required to retrieve the CS of $\bL$ or locate the outlying columns.

Many approaches were developed to address this problem, including\cite{lamport2,lerman2014fast,lamport8,lamport10,lamport13,lamport18,lamport21,lamport24,lamport47,lamport48,zhang2014novel,soltanolkotabi2012geometric,tsakiris2015dual,lerman2015robust,rahmani2016coherence,lamport21}.
In \cite{lamport10}, it is assumed that $\bC$ is a column sparse matrix (i.e., only few data points are outliers) and a matrix decomposition algorithm is proposed to decompose the data into low rank and column-sparse components. In \cite{lamport21}, the $\ell_2$-norm in the PCA optimization problem is replaced with an $\ell_{1,2}$-norm to grant robustness against outliers. In \cite{rahmani2016coherence}, we leveraged the low mutual coherence between the outlier data points and the other data points to set them apart from the inliers. An alternative approach relies on the observation that small subsets of the columns of $\bL$ are linearly dependent (since $\bL$ lies in a low-dimensional subspace) while small subsets of the outlier columns are not given that outliers do not typically follow low-dimensional structures.
Several algorithms exploit this feature to locate the outliers \cite{soltanolkotabi2012geometric,lamport8}.


\subsection{Notation}
Capital and small letters are used to denote matrices and vectors, respectively. For a matrix $\bA$, $\ba_i$ is the $i^{\text{th}}$ row of $\bA$, $\ba^i$ its $i^{\text{th}}$ column,
and $\calN_a$ its null space, i.e., $\calN_a$ is the complement of the row space of $\bA$. $\|\bA\|$ denotes its spectral norm, $\| \bA \|_{*}$ its nuclear norm which is the sum of the singular values, and $\| \bA \|_1$ its $\ell_1$-norm given by $ \| \bA \|_1 = \sum \limits_{i,j} \big | \bA (i,j) \big |$. In addition, the matrix $\bA^{-i}$ denotes the matrix $\bA$ with its $i^{\text{th}}$ column removed.
 In an $N$-dimensional space, $\be_i$ is the $i^{\text{th}}$ vector of
the standard basis. For a given vector $\ba$, $\| \ba \|_p$ denotes its $\ell_p$-norm. The element-wise functions $\sgn (.)$ and $|\: . \: |$ are the sign and absolute value functions, respectively.

\subsection{The identifiability problem}
In this paper, we consider a generalized data model which incorporates simultaneous element-wise and column-wise data corruption. In other words, the given data matrix $\bD\in\mathbb{R}^{N_1\times N_2}$ can be expressed as $\bD = \bL + \bS + \bC$, where $\bL$ is the low rank matrix, $\bC$ contains the outliers, and $\bS$ an element-wise sparse matrix with a arbitrary support. Without loss of generality, we assume that the columns of $\bL$ and $\bS$ corresponding to the non-zero columns of $\bC$ are equal to zero. We seek a robust PCA algorithm that can exactly decompose the given data matrix $\bD$ into $\bL$, $\bS$ and $\bC$. Without any further assumptions, this decomposition problem is clearly ill-posed. Indeed, there are many scenarios where a unique decomposition of $\bD$ may not exist. For instance, the low rank matrix can be element-wise sparse, or the non-zero columns of $\bC$ can be sparse, or the sparse matrix can be low rank. In the following, we briefly discuss various identifiability issues.
\smallbreak
\noindent
\textbf{1.} \emph{Distinguishing $\bL$ from $\bS$}: The identifiability of the low rank plus sparse decomposition problem \cite{lamport22,lamport1} in which $\bD = \bL + \bS$ was studied in \cite{lamport1}.
This problem was shown to admit a unique decomposition as long as the column and row spaces of $\bL$ are sufficiently incoherent with the standard basis
and the non-zero elements of $\bS$ are sufficiently diffused (i.e., not concentrated in few columns/rows)
These conditions are intuitive in that they essentially require the low rank matrix to be non-sparse and the sparse matrix not to be of low rank.

\smallbreak
\noindent
\textbf{2.} \emph{Distinguishing outliers from inliers}: Consider the outlier detection problem in which $\bD = \bL+\bC$. 
Much research was devoted to study different versions of this problem, and various requirements on the distributions of the inliers and outliers were provided to warrant successful detection of outliers.
The authors in \cite{lamport10} considered a scenario where $\bC$ is column-sparse, i.e., only few data columns are actually outliers, and established guarantees for unique decomposition when the rank of $\bL$ and the number of non-zero columns of $\bC$ are sufficiently small.
The approach presented in \cite{soltanolkotabi2012geometric} does not necessitate column sparsity but requires that small sets of outliers be linearly independent. Under this assumption on the distribution of the outliers, exact decomposition can be guaranteed even if a remarkable portion of the data columns are outliers. In this paper, we make the same assumption about the distribution of the outliers, namely, we assume that an outlier cannot be obtained as a linear combination of few other outliers.

\smallbreak
\noindent
\textbf{3.} \emph{Distinguishing a sparse matrix from an outlier matrix:} Suppose $\bD = \bC + \bS$ and assume that the columns of $\bS$ corresponding to the non-zero columns of $\bC$ are equal to zero. Thus, if the columns of $\bS$ are sufficiently sparse and the non-zero columns of $\bC$ are sufficiently dense, one should be able to locate the outlying columns by examining the sparsity of the columns of $\bD$. For example, suppose the support of $\bS$ follows the Bernoulli model with parameter $\rho$ and that the non-zero elements of $\bC$ are sampled from a zero mean normal distribution. If $N_1$ is sufficiently large, the fraction of non-zero elements of a non-zero column of $\bS$ concentrates around $\rho$ while all the elements of a non-zero column of $\bC$ are non-zero with very high probability.


\subsection{Data model}
To the best of our knowledge, this is the first work to account for the simultaneous presence of both sources of corruption. In the numerical examples presented in this paper, we utilize the following data model.

\begin{data model} The given data matrix follows the following model.\\
1. The data matrix $\bD\in\mathbb{R}^{N_1\times N_2}$ can be expressed as
\begin{eqnarray}
\bD = \bL + \bC + \bS \:.
\label{eq:proposedmodel}
\end{eqnarray}
2. $\rank(\bL) = r$. \\
3. Matrix $\bC$ has $K$ non-zero columns. Define $\{ \bg_i \}_{i = 1}^K$ as the non-zero columns of $\bC$. The vectors $\{ {\bg_i} / {\| \bg_i \|} \}_{i = 1}^K$ are i.i.d. random vectors uniformly distributed on the unit sphere $\mathbb{S}^{N_1 - 1}$ in $\mathbb{R}^{N_1}$. Thus, a non-zero column of $\bC$ does not lie in the CS of $\bL$  with overwhelming probability.\\
4. The non-zero elements of $\bS$ follow the Bernoulli model with parameter $\rho$, i.e., each element is non-zero independently with probability $\rho$. \\
5. Without loss of generality, it is assumed that the columns of $\bL$ and $\bS$ corresponding to the non-zero columns of $\bC$, are equal to zero.
\end{data model}
%

\begin{remark}
The uniform distribution of the outlying columns (the non-zero columns of $\bC$) over the unit sphere is not a necessary requirement for the proposed methods. We have made this assumption in the data model to ensure the following requirements are satisfied with high probability:

\begin{itemize}
\item The non-zero columns of $\bC$ do not lie in the CS of $\bL$.

\item The non-zero columns of $\bC$  are not sparse vectors.

\item A small subset of the non-zero columns of $\bC$ is linearly independent.

\end{itemize}
Similarly, the Bernoulli distribution of the non-zero elements of the sparse matrix $\bS$ is not a necessary requirement. This assumption is used here to ensure that the support is not concentrated in some columns/rows. This is needed for $\bS$ to be distinguishable from the outlier matrix $\bC$ and for ensuring that the sparse matrix is not low rank with high probability.
\end{remark}

The proposed data model is pertinent to many applications of machine learning and data analysis. Below, we provide two scenarios motivating the data model in (\ref{eq:proposedmodel}).  

I. Facial images with different illuminations were shown to lie in a low dimensional subspace \cite{lamport44}. Now a given dataset consists of some sparsely corrupted face images along with few images of random objects (e.g., buildings, cars, cities, etc). The images of the random objects cannot be modeled as face images with sparse corruption, which calls for means to recover the face images while being robust to the presence of random images.

II. A users rating matrix in recommender systems can be modeled as a low rank matrix owing to the similarity between people's preferences for different products. To account for natural variability in user profiles, the low rank plus sparse matrix model can better model the data. However, profile-injection attacks, captured by the matrix $\bC$, may introduce outliers in the user rating databases to promote or suppress certain products.
The model (\ref{eq:proposedmodel}) captures both element-wise and column-wise abnormal ratings.

\subsection{Motivating scenarios}
\begin{figure}[t!]
 \centering
    \includegraphics[width=0.35\textwidth]{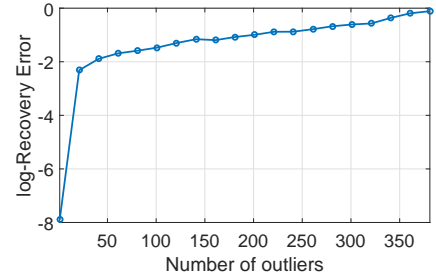}
    \vspace{-0.3cm}
    \caption{The subspace recovery error of (\ref{eq: PCP}) versus the number of outliers.}
    \label{fig:erorpcp1}
\end{figure}

The low rank plus sparse matrix decomposition algorithms -- which only consider the presence of $\bS$ -- are not applicable to our generalized data model given that $\bC$ is not necessarily a sparse matrix. Also, when $\bC$ is column-sparse, it may well be low rank which violates the identifiability conditions of the PCP approach for the low rank plus sparse matrix decomposition \cite{lamport1}.
As an illustrative example, assume $\bD \in \mathbb{R}^{300 \times 500}$ follows Data model 1 with $\rho=0.01$. We apply the decomposition method (\ref{eq: PCP}) to $\bD$ and learn the CS of $\bL$ from the obtained low rank component. Define $$ \text{Log-Recovery Error} =  \log_{10} \left(  \| \bU  -  \hat{\bU} \hat{\bU}^T \bU \|_F / \| \bU \|_F \right) \:,$$ where $\bU$ is an orthonormal basis for the CS of $\bL$ and $\hat{\bU}$ is the learned basis. Fig. \ref{fig:erorpcp1} shows the log recovery error versus $K$. Clearly, (\ref{eq: PCP}) cannot yield correct subspace recovery in the presence of outliers.

\begin{figure}[t!]
 \centering
    \includegraphics[width=0.35\textwidth]{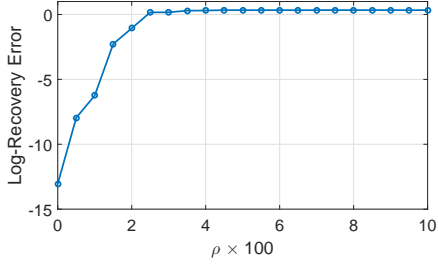}
    \vspace{-0.3cm}
    \caption{The subspace recovery error of the robust PCA algorithm presented in \cite{lerman2014fast} versus $\rho$. }
    \label{fig:erorpcp2}
\end{figure}

On the other hand, robust PCA algorithms that solely consider the column-wise corruption are bound to fail in the presence of the sparse corruption $\bS$ since a crucial requirement of such algorithms is that a set of the columns of $\bD$ lies in the CS of $\bL$.
However, in presence of the sparse corruption matrix $\bS$, even the columns of $\bD$ corresponding to the zero columns of $\bC$ might not lie in the CS of $\bL$. For instance, assume $\bD \in \mathbb{R}^{100 \times 400}$ follows Data model 1 with $K = 200$ and $r = 5$. Fig. \ref{fig:erorpcp2} shows the log recovery error versus $\rho$. In this example, the robust PCA algorithm presented in \cite{lerman2014fast} is utilized for subspace recovery. It is clear that the algorithm cannot yield correct subspace recovery for $\rho \ge 0.01$. The work of this paper is motivated by the preceding shortcomings of existing approaches.

On a first thought, one may be able to tackle the simultaneous presence of sparse corruption and outliers by solving 
\begin{eqnarray}
\begin{aligned}
& \underset{\dot{\bL},\dot{\bS}}{\min}
& & \|\dot{\bL} \|_* + \lambda\|\dot{\bS}\|_1 + \gamma \|\dot{\bC} \|_{1,2} \\
& \text{subject to}
& & \dot{\bL} + \dot{\bS} + \dot{\bC} = \bD \: ,
\end{aligned}
\label{eq:PCPcombine}
\end{eqnarray}
where $\gamma$ and $\lambda$ are regularization parameters. This formulation combines the norms used in the algorithms in \cite{lamport10} and \cite{lamport1}. However, this method requires tuning two parameters but, more importantly, inherits the limitations of
 \cite{lamport10}. Specifically, the PCP approach in \cite{lamport10} requires the rank of $\bL$ to be substantially smaller than the dimension of the data, and fails when there are too many outliers. Also, our experiments have shown that (\ref{eq:PCPcombine}) does not yield an accurate decomposition of the data matrix. For illustration, consider $\bD \in \mathbb{R}^{200 \times 400}$ following Data model 1 with $\rho = 0.01$, and $K = 100$. Let $\bS^{'}$ be the columns of $\bS$ indexed by the complement of the column support of $\bC$
 and $\hat{\bS}^{'}$ the corresponding sparse component recovered by (\ref{eq:PCPcombine}). Table \ref{tabl 2} shows the error in recovery of the sparse component  ${\| \bS^{'} - {\hat{\bS}}^{'} \|_F} / { \| \bS^{'} \|_F }$ versus $r$.
As shown, the convex program in (4) which combines (\ref{eq: PCP}) and \cite{lamport10} cannot yield an accurate decomposition knowing that in the absence of column outliers (i.e., when $K = 0$) and setting $\gamma=0$, (\ref{eq: PCP}) does recover the sparse component with recovery error below 0.01 for all values of $r$ in Table \ref{tabl 2}.

\subsection{Summary of contributions}
In this paper, we develop a new robust PCA approach, dubbed the sparse approximation approach, which can account for both types of corruptions simultaneously.  Below, we provide a summary of contributions.
\begin{itemize}
\item The sparse approximation approach:  In this approach, we put forth an $\ell_1$-norm minimization formulation that allows us to find a sparse approximation for the columns of $\bD$ using a sparse representation. This idea is used to locate the outlying columns of $\bD$ which, once identified, reduces the primary problem to one of a low rank plus sparse matrix decomposition.

\begin{table}
\centering
\caption{Recovery error in the sparse component using algorithm (\ref{eq:PCPcombine}).}
\vspace{-0.25cm}
\begin{tabular}{| c | c | c | c | c |}
\hline
 $r $ & 2 & 5 & 10 & 15 \\
 \hline
 ${\| \bS^{'} - {\hat{\bS}}^{'} \|_F} / { \| \bS^{'} \|_F }$ & 0.04
 & 0.26
 & 0.51
 & 0.65 \\
 \hline
\end{tabular}
\label{tabl 2}
\vspace{-.25cm}
\end{table}

\item We develop a new randomized design that provides a scalable implementation of the proposed method. In this design, the CS of $\bL$ is learned using few randomly sampled data columns. Subsequently, the outliers are located using few randomly sampled rows of the data.

\item We provide a mathematical analysis of the sparse approximation idea underlying our approach where we prove that the $\ell_1$-norm minimization can yield the linear representation of a sparsely corrupted inlier.

\end{itemize}

\subsection{Paper organization}
The rest of the paper is organized as follows. In section \ref{sec:SAtheory}, the idea of sparse approximation is explained. Section \ref{sec:proposemethod} presents the proposed robust PCA method and Section \ref{sec:numerical} exhibits the numerical experiments. The proofs of all the theoretical results are provided in the appendix along with additional theoretical investigations of the sparse approximation problem.

\section{Sparse approximation of sparsely corrupted data}
\label{sec:SAtheory}
Suppose the vector $\bh$ lies in the CS of a matrix $\bH$, i.e., $\bh = \bH \bb$, where vector $\bb$ is the linear representation of $\bh$ with respect to the columns of $\bH$. In a least-square sense, this representation can be obtained as the optimal point of $\underset{{\hat{\bb}}}{\min} \:\:
 \|  \bH \hat{\bb} - \bh \|_2$. The main question that this section seeks to address is whether we can recover such a representation using a convex optimization formulation when both $\bH$ and $\bh$ are sparsely corrupted.
In this section, we propose an $\ell_1$-norm minimization problem and prove that it can yield the underlying representation. We refer to this approach as ``sparse approximation''.

\subsection{Theoretical result}
The following definition presents 
the notion of sparse approximation. 

%
\begin{definition}
\label{def:sparse_approx}
Suppose $\bA \in \mathbb{R}^{N_1\times n}$ can be expressed as $\bA = \bB+\bS$, where $\bS$ is a sparse matrix, $n \ll N_1$, and $r_b < n$ ($r_b$ is the rank of $\bB$). We say that $\bA^{-i} \bz$, where $\bz \in \mathbb{R}^{n-1}$, is a sparse approximation of $\ba^{i}$ if $\bb^i = {\bB}^{-i}\bz$.
Thus, if
 $\bA^{-i} \bz$ is a sparse approximation of $\ba^{i}$, then
$\bA^{-i}\bz - \ba^i = {\bS}^{-i}\bz - \bs^i$.
\end{definition}

\noindent
The reason that we refer to $\bA^{-i} \bz$ as the sparse approximation of $\ba^{-i}$ is that $\bA^{-i}\bz - \ba^i = {\bS}^{-i}\bz - \bs^i$ is a sparse vector if $\rho$ and $n$ are sufficiently small.

Assume that the CS of $\bB$ does not include sparse vectors, i.e., the CS of $\bB$ is not coherent with the standard basis.
According to Definition \ref{def:sparse_approx}, if $\rho$ and $n$ are small enough and $\bb^i$ lies in the CS of ${\bB}^{-i}$, then $\bA^{-i} \bz^{*}$ is a sparse approximation of $\ba^i$, where $\bz^{*}$ is the optimal point of
\begin{eqnarray}
\underset{{\bz}}{\min} \:\:
 \|  \bA^{-i} \bz - \ba^i \|_0 \:,
\label{eq:introduction_ell0}
\end{eqnarray}
because the span of $\bB$ does not contain sparse vectors so the only way to obtain a sparse linear combination is to cancel out $\bb^i$. 
 Moreover, we could show that if the $\ell_0$-norm of (\ref{eq:introduction_ell0}) is relaxed to an $\ell_1$-norm, we will still able to obtain the sparse approximation.
The following lemma establishes that the sparse approximation can be recovered by solving a convex $\ell_1$-norm minimization problem if $\bS$ is sufficiently sparse. In order to obtain concise sufficient conditions, Lemma 1 assumes a randomized model for the distribution of the rows of $\bB$ in its row space. In the appendix, we present deterministic sufficient conditions for a more general optimization problem.

\begin{assumption}
The rows of the matrix $\bB$ are i.i.d. random vectors uniformly
distributed on the intersection of the row space of $\bB$ and the unit sphere $\calS^{n-1}$. 
\label{amp:random}
\end{assumption}

\noindent
Before we state the lemma, we  define $\bz_o^{*}$ as the optimal point of the following oracle optimization problem
\begin{eqnarray}
\underset{{\bz}}{\min} \:\:
 \|  \bS^{-i} \bz - \bs^i \|_1  \quad \text{s.t.} \quad \bB^{-i} \bz = \bb^i \: .
 \label{eq:oracle_optz}
\end{eqnarray}
In addition, define $n_s$ and $n_s^{'}$ as the cardinalities of $\calL_S^c$ and $\calL_{\bz_o^{*}} \cap \calL_S^c$, respectively, where $\calL_S$ and $\calL_{\bz_o^{*}}$ are defined as
\begin{eqnarray}
\begin{aligned}
& {\cal L}_S = \{k \in [N_1] : \bs_{k} = 0 \} \:, \\
&\calL_{\bz_o^{*}} = \{k \in [N_1]  : (\bs_{k})^T \bz_o^{*} = 0 \} \:.
\end{aligned}
\label{eq:set_definition}
\end{eqnarray}
Thus, $n_s$ is the number of non-zero rows of $\bS$ and $n_s^{'}$  is the number of non-zero rows of $\bS$ which are orthogonal to $\bz_o^{*}$. If
 the support of $\bS$ follows the random model,  $n_s^{'}$ is much smaller than $n_s$. In addition, if $n \ll N_1$ and $\rho$ is small enough, $n_s$ will be much smaller than $N_1$.

\begin{lemma}
Suppose matrix $\bA \in \mathbb{R}^{N_1 \times n}$ is a full rank matrix which can be expressed as $\bA = \bB+\bS$, where $\bB$ follows Assumption \ref{amp:random}. Define
\begin{eqnarray}
\nonumber
\xi = \sqrt{ \frac{2}{\pi}} \frac{N_1 - n_s}{\sqrt{r_b}}  - 2 \sqrt{N_1 - n_s} - t_1 \sqrt{\frac{N_1 - n_s}{r_b -1}} \:,
\end{eqnarray}
and define $\bR_{b}$ and $\bP_{b}$ as orthonormal bases for the row space and null space of $\bB$, respectively. If
\begin{eqnarray}
\begin{aligned}
& \frac{1}{2} \: \xi  > n_s^{'} + \sum_{i \in \calL_{\bz_o^{*}}} \|\bs_i \| + \sqrt{n_s - n_s^{'}}t_2  \quad \text{and}  \\
& \frac{ \| \be_i^T \bP_{b} \|}{ 2 \| \be_i^T \bR_{b}\|} \: \xi > \sum_{i \in \calL_{\bz_o^{*}}} \|\bs_i \| + \sqrt{n_s - n_s^{'}}t_2 \:,
\end{aligned}
\label{eq:suff_randomi}
\end{eqnarray}
then $\bz_o^{*}$ (the optimal point of (\ref{eq:oracle_optz})) is the optimal point of
\begin{eqnarray}
\underset{{\bz}}{\min} \:\:
 \|  \bA^{-i} \bz - \ba^i \|_1 \:
\label{eq:introduction_el1}
\end{eqnarray}
with probability at least $1 - \exp(- t_1^2 / 2) - \exp \left(-\frac{r_b}{2} (t_2^2 - \log t_2^2 - 1) \right)$, for all $t_1 \ge 0 \: , \: t_2 >1 $. 
\label{lm:yek}
\end{lemma}

\begin{remark}
Suppose the support of $\bS$ follows the Bernoulli model with parameter $\rho$.
If $\rho$ and $n$ are small enough, $n_s \ll N_1$. Thus, the order of $\xi$ is roughly $\frac{N_1}{\sqrt{r_b}}$. Define $\kappa$ such that $\kappa \ge \underset{i,j}{\max} \: {\bS (i,j)}$. The vector $\bz_o^{*}$ cannot be simultaneously orthogonal to too many non-zero rows of $\bS$. Thus, $n_s^{'}$ is much smaller than $n_s$. Therefore, the order of the RHS of (\ref{eq:suff_randomi}) is roughly $\calO \Big( \sqrt{n_s} +  n_s^{'} \kappa ( \sqrt{\rho n }+1) \Big)$. If we assume that the row space of $\bB$ is a random $r_b$-dimensional subspace in the $n$-dimensional space, then $$\mathbb{E} \left\{ \frac{ \| \be_i^T \bP_b \|}{  \| \be_i^T \bR_b \|} \right\} = \frac{n - r_b}{r_b}\:.$$
Accordingly, the sufficient conditions in Lemma \ref{lm:yek} amount to a requirement that $\frac{(n - r_b) N_1}{r_b \sqrt{r_b} }$ is sufficiently larger than $ \Big( \sqrt{n_s} +  n_s^{'} \kappa ( \sqrt{\rho n }+1) \Big)$. In our problem, $r_b \ll N_1$ and $n_s \ll N_1$. Thus, the sufficient conditions of Lemma \ref{lm:yek} are naturally satisfied.
\end{remark}

\begin{algorithm}
\caption{Sparse Approximation Approach}
{\footnotesize
\textbf{Input}: Data matrix $\bD \in \mathbb{R}^{N_1 \times N_2} $

\smallbreak
\textbf{1. Outlying Columns Detection}\\
\textbf{1.1} Define $\{ \bz_i^{*} \}_{i=1}^{N_2}$ as the optimal points of
\begin{eqnarray}
\begin{aligned}
\underset{{\bz}}{\min} \:\:
 \|  \bD \bz \|_1 + \lambda  \| \bz \|_1
\quad \text{s. t.}  \quad    \bz^T \be_i  = 1
\end{aligned}
\label{eq:convex_final}
\end{eqnarray}
for $\{ \be_i \}_{i = 1}^{N_2}$, respectively. If $\bD \bz^{*}_k$ is not sufficiently sparse, it is concluded that the $k$-th column of $\bD$ is an outlying column. Form set $\calI_o$ as the index set of detected outlying columns.\\
\textbf{1.2} Form matrix $\bM$ which is equal to $\bD$ with the detected outlying columns removed.

\smallbreak
\textbf{2. Matrix Decomposition}\\
Obtain $\bL_m$ and $\bS_m$ as the optimal point of
\begin{eqnarray}
\begin{aligned}
& \underset{\dot{\bL}_m,\dot{\bS}_m}{\min}
& &  \frac{1}{\sqrt{N_1}} \|\dot{\bS}_m\|_1  + \|\dot{\bL}_m \|_* \\
& \text{subject to}
& & \dot{\bL}_m + \dot{\bS}_m = \bM \: \\
\end{aligned}
\end{eqnarray}

\smallbreak
\textbf{Output:} Set $\calI_o$ as the identified outlying columns and $\bL_m$ and $\bS_m$ as the low rank and sparse components of the non-outlying columns of $\bD$.
 }
\end{algorithm}

\section{Proposed sparse approximation method}
\label{sec:proposemethod}
In this section, the sparse approximation (SA) method is presented. We also present a randomized design which can reduce the complexity of the proposed method from $\calO(N_2^3)$ to $\calO (r^2 \max(N_1 , N_2))$.
The table of Algorithm 1 presents the proposed algorithm. 
The $\ell_1$-norm functions are utilized to enforce sparsity to both the representation vector $\bz$ and the residual vector $\bD \bz$ \cite{lamport25,lamport7,decod,candes2007sparsity,joneidi2015matrix}.
The main idea is to first locate the non-zero columns of $\bC$ to reduce the problem to that of a low rank plus sparse matrix decomposition. In order to identify the outliers, we attempt to find a sparse approximation for each data column $\bd^i, i = 1,\ldots,N_2$, using a sparse linear combination of the columns of $\bD^{-i}$.  
If for certain columns such an approximation cannot be found, we identify these columns as outliers.

\begin{figure}[t!]
 \centering
    \includegraphics[width=0.50\textwidth]{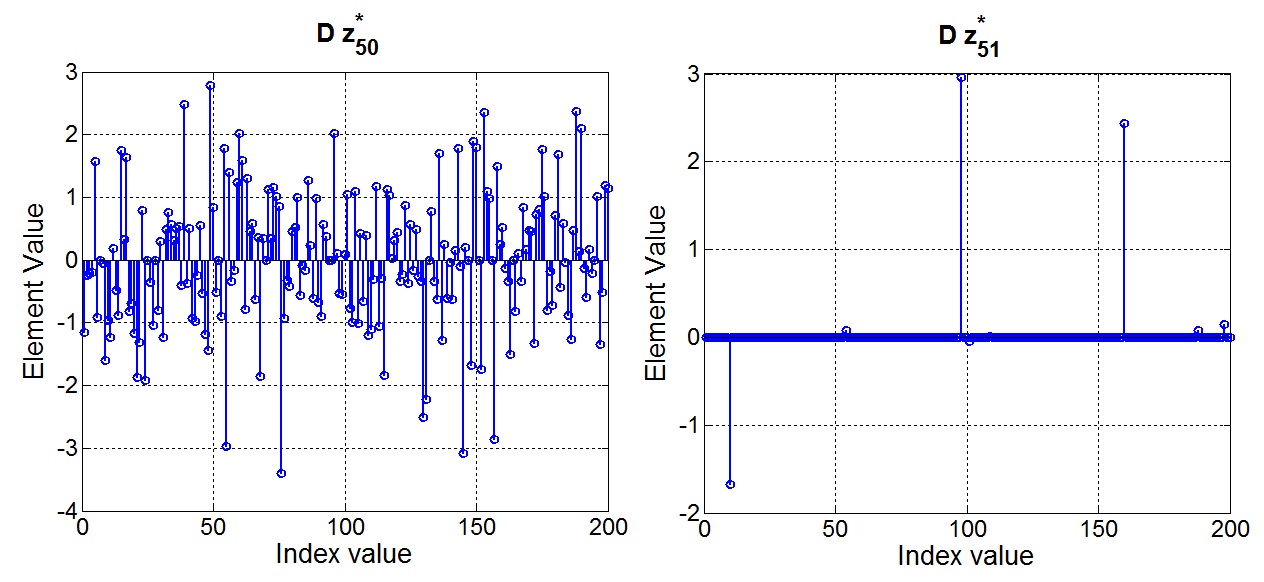}
    \vspace{-0.3cm}
    \caption{The elements value of $\bD \bz_{50}^{*}$ and $\bD \bz_{51}^{*}$. The $50^{\text{th}}$ column is an outlying column and the $51^{\text{th}}$ column is not an outlying column.}
    \label{fig:example 1}
\end{figure}

The key idea underlying this approach is that the sparsity rate of $\bD\bz_i^*$ in (\ref{eq:convex_final}) can be used to certify the identity of $\bd^i$ being an outlier or a sparsely corrupted inlier. Before providing some insight, we consider an illustrative example in which $\bD \in \mathbb{R}^{200 \times 500}$ follows Data model 1, $r = 5$, $\rho = 0.01$ and the first 50 columns of $\bC$ are non-zero ($K = 50$). Fig. \ref{fig:example 1} shows $\bD \: \bz_{50}^{*}$ and $\bD \: \bz_{51}^{*}$. The outlying column is clearly distinguishable.
\smallbreak
\noindent\emph{Insight for the SA method:} To gain more insight, consider the scenario where the $i$-th column of $\bC$ is zero, i.e., $\bc^i = \mathbf{0}$,  so that the $i$-th data column is a sparsely corrupted inlier. In this case, if the regularization parameter $\lambda$ is chosen appropriately, (\ref{eq:convex_final}) can identify a sparse vector $\bz_i^*$ (sparsity of $\bz$ is promoted by the $\ell_1$-norm regularizer) such that $\bD\bz_i^*$ is also sparse.
The $\ell_1$-norm  functions forces (\ref{eq:convex_final}) to put the non-zero values of $\bz_i^{*}$ on the columns of $\bD$ such that a linear combination of their low rank components cancel out the low rank component of $\bd^i$ (i.e. they provide a sparse approximation for $\bd_i$) and the linear combination of their sparse component yields a sparse vector. In other word, $\bL \bz_i^{*} = 0$ and $\bS \bz_i^{*}$ is a sparse vector since it is a linear combination of few sparse vectors and the algorithm automatically puts the non-zero values of $\bz_i^{*}$ on the columns such that $\bS \bz_i^{*}$ is roughly as sparse as possible.

On the other hand, if $\bc^i \ne \mathbf{0}$, i.e., $\bc^i$ is an outlying column, $\bD\bz_i^*$ is not likely to be sparse for a sparse $\bz_i^*$ since small subsets of outlying columns are linearly independent, and an outlying column is unlikely to admit a sparse representation in the sparsely corrupted columns of $\bL$, to say that linear combinations of few sparsely corrupted columns of $\bL$ are unlikely to approximate an outlying column.

\subsection{Randomized implementation of the proposed method}
Randomized techniques are utilized to reduce the sample and computational complexities of robust PCA algorithms \cite{liu2011solving,myp,lamport27,lamport28,lamport43,vay,zhou2011godec,rahmani2015randomized}.
Algorithm 1 solves an $N_2 \times N_2$ dimensional optimization problem to identify all the outlying columns. However, here we show that this problem can be simplified to a low-dimensional subspace learning problem. Let $\bL = \bU \mathbf{\Sigma} \bV^T$ be the compact singular value decomposition (SVD) of $\bL$, where $\bU \in \mathbb{R}^{N_1 \times r}$, $\mathbf{\Sigma} \in \mathbb{R}^{r \times r}$ and $\bV \in \mathbb{R}^{N_2 \times r}$.
We can rewrite (\ref{eq:proposedmodel}) as
\begin{eqnarray}
\bD = \bU \bQ  + \bS + \bC \:,
\end{eqnarray}
where $\bQ = \mathbf{\Sigma}\bV^T$. We call $\bQ$ the representation matrix.
The table of Algorithm 2 details the randomized implementation of the proposed SA method. In the randomized implementation, first the CS of $\bL$ is obtained using a random subset of the data columns. Matrix $\bD_{ \phi_1 } \in \mathbb{R}^{N_1 \times m_1}$ consists of $m_1$ randomly sampled columns. The proposed outlier detection approach is applied to $\bD_{ \phi_1 }$ to identify the outlying columns of $\bD_{ \phi_1 }$. The matrix $\bD_{ \phi_1 }$ can be expressed as
\begin{eqnarray}
\bD_{ \phi_1 } = \bL_{ \phi_1 }  + \bS_{ \phi_1 } + \bC_{ \phi_1 } \:,
\end{eqnarray}
where $\bL_{ \phi_1 }$, $\bS_{ \phi_1 }$ and $\bC_{ \phi_1 }$ are the corresponding columns sampled from $\bL$, $\bS$ and $\bC$, respectively. If $m_1$ is sufficiently large, $\bL_{ \phi_1 }$ and $\bL$ will have the same CS. Thus, if we remove the outlying columns of $\bD_{ \phi_1 }$ and decompose the resulting matrix, $\bM_{ \phi_1 }$, the obtained low rank component can yield the CS of $\bL$.

Suppose the CS of $\bL$ is learned correctly and assume that the $i$-th column of $\bC$ is equal to zero. Thus, the $i$-th column of $\bD$ can be represented as
\begin{eqnarray}
\bd^i = \bU \bq^i + \bs^i \:.
\end{eqnarray}
It was shown in \cite{myp} that $\bq^i$ can be obtained as the optimal point of
\begin{eqnarray}
\underset{\dot{\bq}}{\min} \| \mathbf{\Phi}_2^T \bd^i - \mathbf{\Phi}_2^T \bU \dot{\bq}\|_1 \:,
\label{eq rep learn 1}
\end{eqnarray}
if $m_2$ is sufficiently large and some mild sufficient conditions are satisfied.
Define $\bq^{*}$ as the optimal point of (\ref{eq rep learn 1}). If $\bq^{*} = \bq^i$, then $$\mathbf{\Phi}_2^T \bd^i - \mathbf{\Phi}_2^T \bU {\bq}^{*} = \mathbf{\Phi}_2^T \bs^i \: .$$
The vector $\mathbf{\Phi}_2^T \bs^i$ consists of $m_2$ randomly sampled elements of the sparse vector $\bs^i$. Accordingly, if $\bc^i$ (the $i$-th column of $\bC$) is equal to zero, then $\mathbf{\Phi}_2^T \bd^i - \mathbf{\Phi}_2^T \bU {\bq}^{*}$ is a sparse vector. However, if $\bc^i$ is not equal to zero and $m_2$ is sufficiently large, it will be highly unlikely that $\mathbf{\Phi}_2^T \bd^i - \mathbf{\Phi}_2^T \bU {\bq}^{*} $ is a sparse vector since $\mathbf{\Phi}_2^T \bU {\bq}^{*}$ cannot cancel out the component of $\mathbf{\Phi}_2^T \bc^i$ that does not lie in the CS of $\mathbf{\Phi}_2^T \bU$.

\begin{remark}
In the CS learning step, we identify the outlying columns via the sparsity of $\bD_{ \phi_1 } \bz_i^{*}$. If $\bz_i^{*}$ lies in the null space of $\bL_{\phi_1}$, then $\bz_i^{*}$ has $c \: r$ non-zero elements where $c > 1$. According to our investigations, if $\lambda$ is chosen appropriately, $c$ is a small number mostly smaller than 3. Thus, $\bD \bz_i^{*}$ has at most $2 r \rho N_1$ non-zero elements. In practice, it is much smaller than $2 r \rho N_1$ since the optimization searches for the most sparse linear combination. In step 3 of the algorithm, the outlying columns are located by examining the sparsity of the columns of $\bH = \bD_{ \phi_2 } - \mathbf{\Phi}_2^T \hat{\bU} \hat{\bQ}$. If the $i$-th column of $\bC$ is equal to zero, $\hat{\bU}$ and $\hat{\bQ}$ are correctly recovered, and $m_2$ is sufficiently large, then the $i$-th column of $\bH$ is a sparse vector with roughly $\rho m_2$ non-zero elements. Accordingly, if we set an appropriate threshold for the number of dominant non-zero elements, the outlying columns are correctly identified.
\end{remark}

\begin{algorithm}
\caption{Randomized Implementation of the Sparse Approximation Method}
{\footnotesize
\textbf{Input}: Data matrix $\bD \in \mathbb{R}^{N_1 \times N_2} $\\
\textbf{1. Initialization}: Form the column sampling matrix $\mathbf{\Phi}_1 \in \mathbb{R}^{N_2 \times m_1} $ and the row sampling matrix $\mathbf{\Phi}_2 \in \mathbb{R}^{N_1 \times m_2} $.\\
\textbf{2. CS Learning}\\
\textbf{2.1} Column sampling: The matrix $\mathbf{\Phi}_1$ samples $m_1$ columns of the given data matrix, $\bD_{ \phi_1 } = \bD \mathbf{\Phi}_1$. \\
\text{2.1} Sampled outlying columns detection: Define $\{ \bz_i^{*} \}_{i = 1}^{m_1}$ as the optimal point of
\begin{eqnarray}
\underset{{\bz}}{\min} \| \bD_{ \phi_1 } {\bz} \|_1 + \lambda \| \bz \|_1 \quad \text{s.t.} \quad {\bz}^T \be_i = 1 ,
\end{eqnarray}
for $\{  \be_i \}_{i = 1}^{m_1}$. If  $\bD_{ \phi_1 } \bz_k^{*}$ is not a sparse vector, the $k$-th column of $\bD_{ \phi_1 }$ is identified as an outlying column. \\
\textbf{2.2} Obtain $\hat{\bL}_{ \phi_1 }$ and $\hat{\bS}_{ \phi_1 }$ as the optimal point of
\begin{eqnarray}
\begin{aligned}
& \underset{\dot{\bL}_{ \phi_1 },\dot{\bS}_{ \phi_1 }}{\min}
& &  \frac{1}{\sqrt{N_1}} \|\dot{\bS}_{ \phi_1 }\|_1  + \|\dot{\bL}_{ \phi_1 } \|_* \\
& \text{subject to}
& & \dot{\bL}_{ \phi_1 } + \dot{\bS}_{ \phi_1 } = \bM_{ \phi_1 } \:, \\
\end{aligned}
\end{eqnarray}
where $\bM_{ \phi_1 }$ is equal to $\bD_{ \phi_1 }$ with its outlying columns removed. \\
\textbf{2.3} CS recovery: Form the orthonormal matrix $\hat{\bU}$ as a basis for the CS of $\hat{\bL}_{ \phi_1 }$. \\
\textbf{3. Learning $\bQ$ and Locating the Outlying Columns.}\\
\textbf{3.1} Row sampling: The matrix $\mathbf{\Phi}_2$ samples $m_2$ rows of the given data matrix, $\bD_{ \phi_2 } = \mathbf{\Phi}_2^T \bD$.\\

\textbf{3.2} Learning $\bQ$: Obtain $\hat{\bQ}$ as the optimal point of
\begin{eqnarray}
\underset{\dot{\bQ}}{\min} \| \bD_{ \phi_2 } - \mathbf{\Phi}_2^T \hat{\bU} \dot{\bQ}\|_1.
\label{eq: rep learn}
\end{eqnarray}
\textbf{3.3} Outlying column Detection: Form set $\calI_o$ as the index set of the non-sparse columns of $\bD_{ \phi_2 } - \mathbf{\Phi}_2^T \hat{\bU} \hat{\bQ}$. \\
\textbf{4. Obtaining the Low Rank and Sparse Components.}\\ Form $\hat{\bL} = \hat{\bU} \hat{\bQ}$ with the columns indexed by $\calI_o$ set equal to zero. Form $\hat{\bS}$ equal to $\bD - \hat{\bU} \hat{\bQ}$ with its columns indexed by $\calI_o$ set to zero. \\
\textbf{Output:} The matrices $\hat{\bL}$ and $\hat{\bS}$ are the obtained low rank and sparse components, respectively. The set $\calI_o$ contains the indices of the identified outlying columns.
}
\end{algorithm}

\section{Numerical Simulations}
\label{sec:numerical}
In this section, we present a set of numerical experiments to study the performance of the proposed approach. First, we validate the idea of sparse approximation for outlier detection and study its requirements.
Second, we provide a set of phase transition plots to demonstrate the requirements of the randomized implementation, i.e., the sufficient number of randomly sampled columns/rows. Finally, we study the sparse approximation approach for outlier detection with real world data.

\subsection{The idea of outlier detection}
\label{sec: simul_A}
\begin{figure}[t!]
 \centering
    \includegraphics[width=0.30\textwidth]{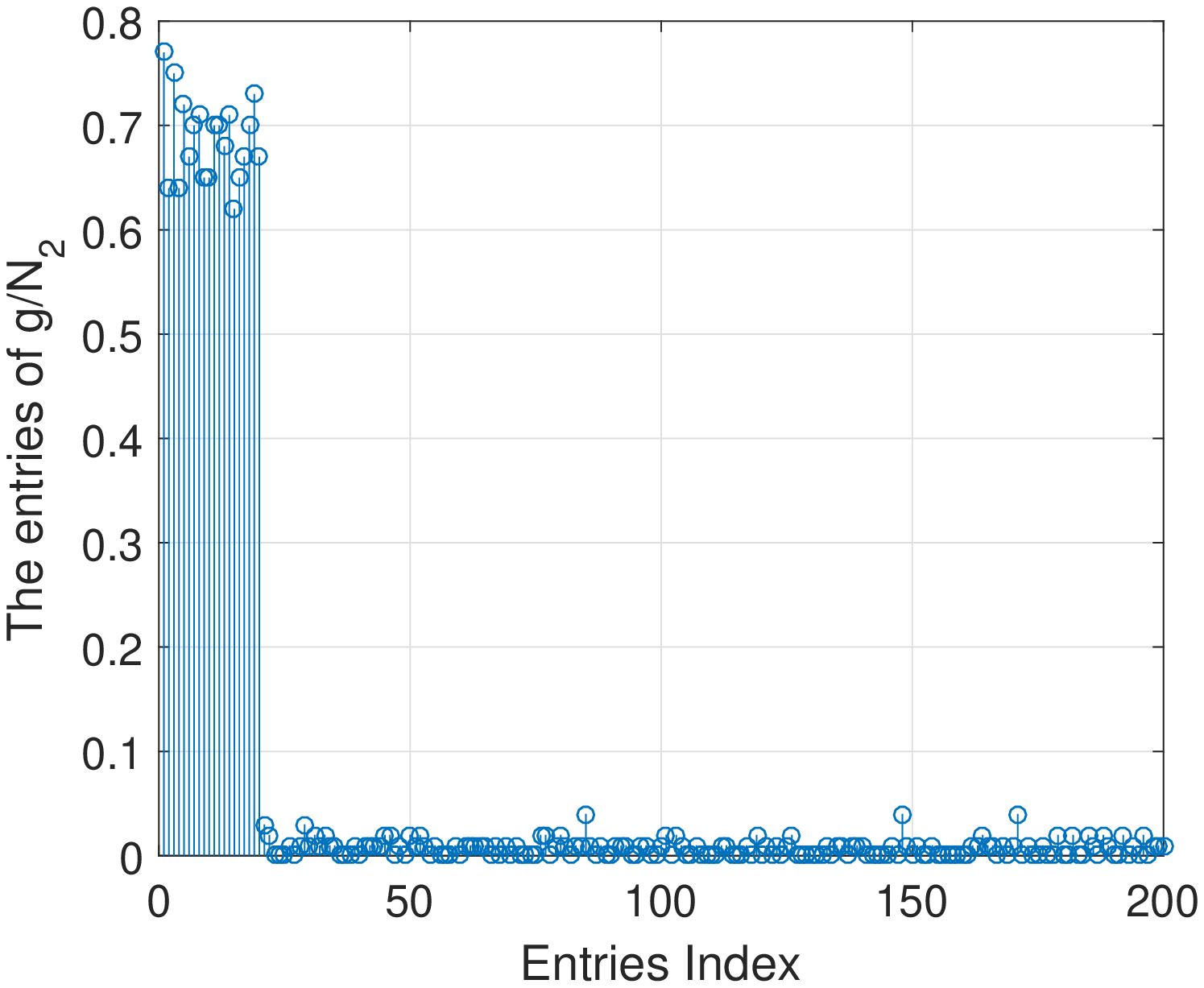}
    \vspace{-0.3cm}
    \caption{The entries of vector $\bg$.  }
    \label{fig:simul1_out}
\end{figure}

Suppose the given data follows Data model 1 and $\bD \in \mathbb{R}^{100 \times 200}$. The first 20 columns of $\bC$ are non-zero. The matrix $\bS$ follows the Bernoulli model with $\rho = 0.01$. The rank of $\bL$ is equal to 5. We solve (\ref{eq:convex_final}) with the constraint vector set equal to $\{ \be_i \}_{i = 1}^{N_2}$ and define $\{ \bz_i^{*} \}_{i = 1}^{N_2}$ as the corresponding optimal point. Define $\bh_i = | \bD \bz^{*}_i | / \max (| \bD \bz^{*}_i | )$. We also, define a vector $\bg \in \mathbb{R}^{N_2 \times 1}$ whose $i$-th entry is set equal to the number elements of $\bh_i$ greater than $0.1$. Thus, the $i$-th element of $\bg$ is the number of dominant non-zero elements of $\bD \bz_i^{*}$. Fig. \ref{fig:simul1_out} shows the elements of $\bg / N_2$. As shown, the indices corresponding to the outlying columns are clearly distinguishable.

\subsection{Phase transition}
\begin{figure}[t!]
 \centering
    \includegraphics[width=0.30\textwidth]{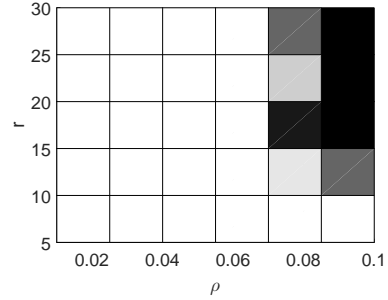}
    \vspace{-0.3cm}
    \caption{Phase transition of the outlier detector in the plane of $r$ and $\rho$.}
    \label{fig:phase}
\end{figure}

In the presented theoretical analysis, we have shown that if the rank of the low rank component is sufficiently small and the sparse component is sufficiently sparse, the $\ell_1$-norm optimization problem can yield the sparse approximation (Lemma \ref{lm:yek} and Theorem \ref{lm: two_ind}). In this section, we assume the data $\bD \in \mathbb{R}^{400 \times 400}$ follows Data model 1 and study the phase transition of Algorithm 1 (which uses sparse approximation for outlier detection) in the 2D-plane of $r$ and $\rho$. The first 200 columns of the given data are outlying columns, i.e., $K = 200$. We define a vector $\bh_i = | \bD \bz^{*}_i | / \max (| \bD \bz^{*}_i | )$ as before corresponding to each data column, and we classify the $i$-th column as an outlier if more than 40 percent of the elements of $\bh_i$ are greater than 0.1. Fig. \ref{fig:phase} shows the phase transition in the plane of $r$ and $\rho$. For each pair $(r,\rho)$, we generate 10 random realizations. In this figure, white designates that all outliers are detected correctly and no inlier is misclassified as an outlier. One can observe that if $\rho < 0.07$, we can correctly identify all the outliers even with $r = 30$.

\begin{figure}[t!]
 \centering
    \includegraphics[width=0.50\textwidth]{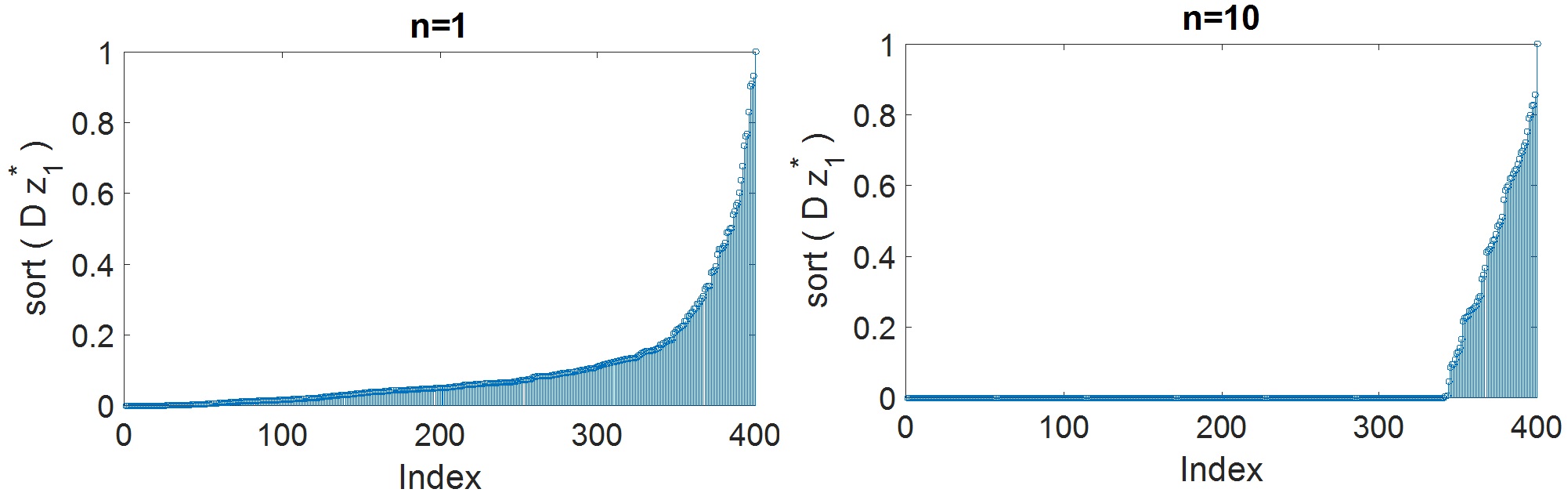}
    \vspace{-0.3cm}
    \caption{The entries of vector $\bD \bz_1^{*}$ for different number of clusters of the columns of $\bL$. In the right plot, the columns of $\bL$ lie in a union of $10$ 1-dimensional subspaces. In the left plot, the columns of $\bL$ lie in one 10-dimensional subspace.}
    \label{fig:union_sorted}
\end{figure}

In practice, the proposed method can handle larger values of $r$ and higher sparsity levels (i.e., more non-zero elements) because the columns of the low rank matrix typically exhibit additional structures such as clustering structures \cite{lamport7,lamport17,rahmani2015innovation}. In the simulation corresponding to Fig. \ref{fig:phase}, the low rank matrix was generated as $\bL = \bU \bQ$, where the elements of $\bU \in \mathbb{R}^{N_1 \times r}$ and $\bQ \in \mathbb{R}^{r \times N_2}$ are drawn from a zero mean normal distribution. Thus, the columns of $\bL$ are distributed randomly in the CS of $\bL$. Accordingly, the elements of $\bz_i^{*}$ must have at least $r+1$ non-zero elements to yield the sparse approximation. However, if the columns of $\bL$ lie in a union of, say, $n$ $r/n$-dimensional subspaces, $r/n + 1$ non-zero elements can be sufficient. Thus, if the data exhibits a clustering structure, the algorithm can bear with higher rank and sparsity levels. As an example, assume $\bD \in \mathbb{R}^{400 \times 120}$ follows Data model 1 with $r = 10$,  $\rho = 0.1$, and $K = 40$ (the last 40 columns are outlying columns).
Fig. \ref{fig:union_sorted} shows the sorted elements of $\bD \bz_1^{*}$. In the left plot, the columns of $\bL$ lie in one 10-dimensional subspace, whereas in the right plot, the columns of $\bL$ lie in a union of ten 1-dimensional subspaces. As shown, the proposed method yields a better output if the data admits a clustering structure.

\begin{figure}[t!]
	\centering
    \includegraphics[width=0.5\textwidth]{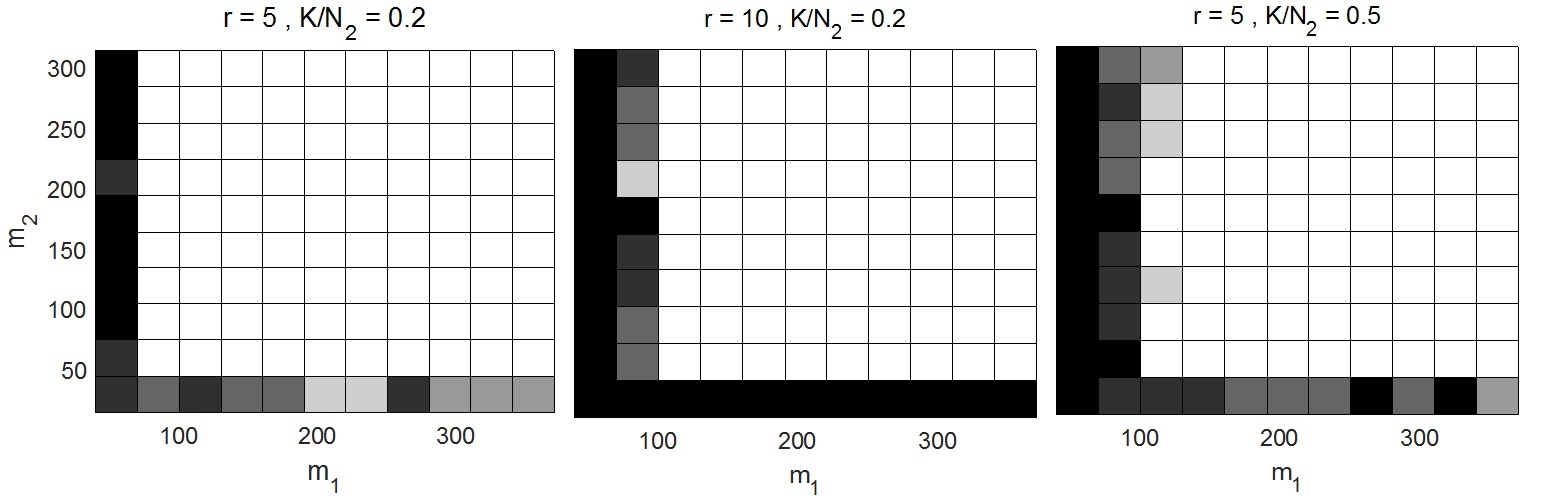}
    \vspace{-.1in}
    \caption{The phase transition plots of Algorithm 2 versus $m_1$ and $m_2$ for different values of $r$ and $K/N_2$. }
    \label{fig:phaserandomized 1}
\end{figure}

\begin{figure}[t!]
	\centering
    \includegraphics[width=0.5\textwidth]{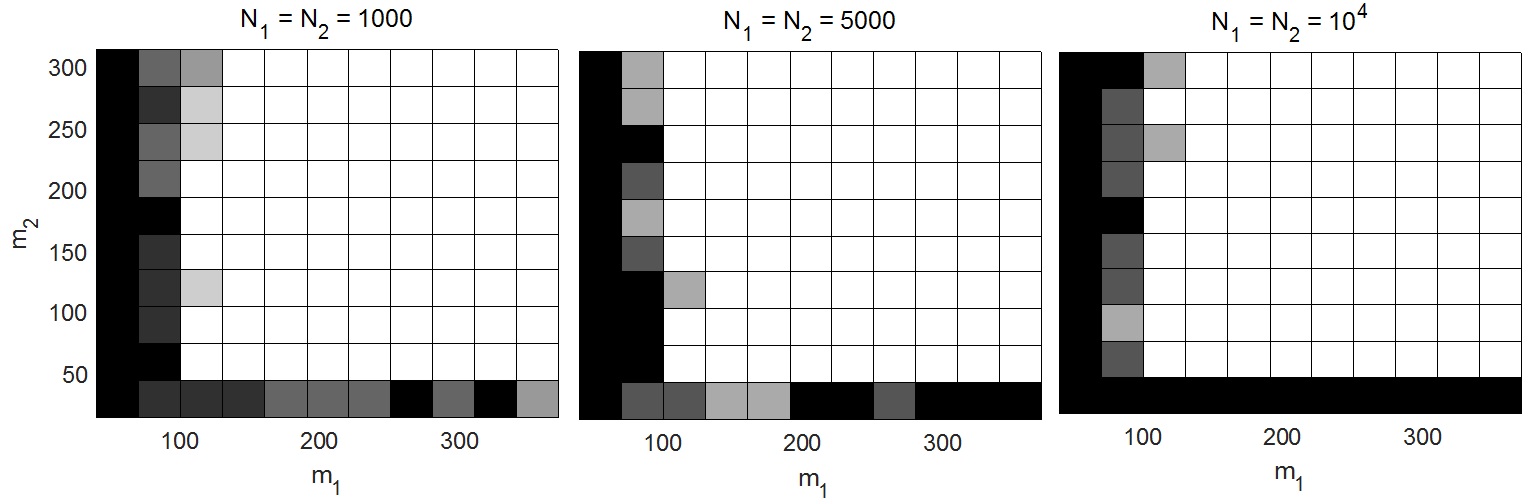}
    \vspace{-.1in}
    \caption{The phase transition plots of Algorithm 2 versus $m_1$ and $m_2$ for different sizes of given data matrix. }
    \label{fig:phaserandomized 2}
\end{figure}

\begin{figure}[t!]
 \centering
    \includegraphics[width=0.3\textwidth]{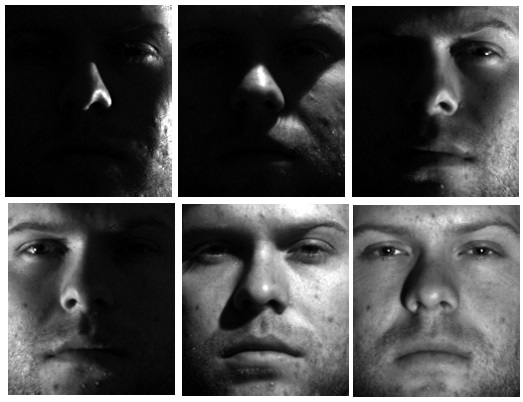}
    \vspace{-0.3cm}
    \caption{Few samples of the face images with different illuminations.}
    \label{fig:face}
\end{figure}

\begin{figure}[t!]
 \centering
    \includegraphics[width=0.4\textwidth]{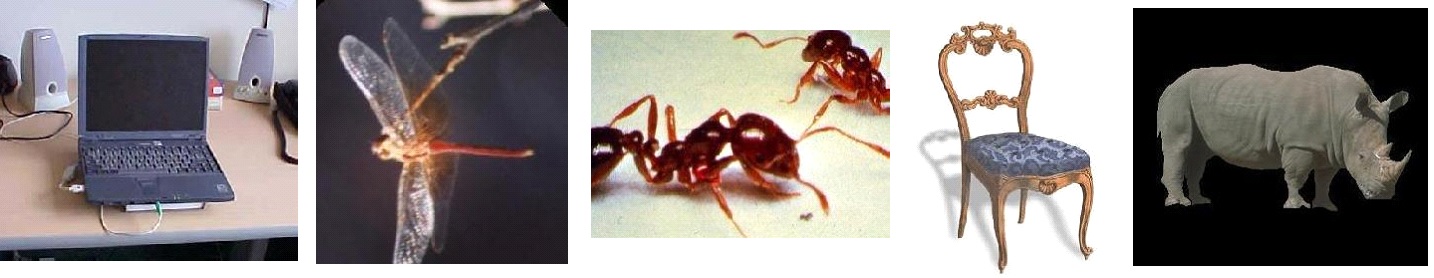}
    \vspace{-0.3cm}
    \caption{Random examples of the images in Caltech101 database.}
    \label{fig:random 101}
\end{figure}

\subsection{Phase transition for the randomized implementation}
In this section, the requirements of Algorithm 2 are studied. The data matrix $\bD \in \mathbb{R}^{1000 \times 1000}$ follows Data model 1. The phase transition shows the performance of Algorithm 2 as function of $m_1$ and $m_2$. A trial is considered successful if the rank of $\hat{\bU}$ is equal to $r$, step 3.3 identifies the outlying columns correctly, and
\begin{eqnarray}
\| (\bI - \bU \bU^T) \hat{\bU} \|_F \leq 10^{-3} \:.
\end{eqnarray}
Fig. \ref{fig:phaserandomized 1} shows the phase transition plots with different values of $r$, $\rho$ and $K/N_2$. One can see that the required values for $m_1$ and $m_2$ increase if the $r$ or $K/N_2$ increase. The required value for $m_1$ is approximately linear in $r \frac{ N_2}{N_2 - K}$ and the required value for $m_2$ is linear in $r$ \cite{myp,rahmani2015randomized}. Fig. \ref{fig:phaserandomized 2}
Shows the phase transition for different dimensions of $\bD$, where $r= 5$, $\rho = 0.02$ and $\frac{K}{N_2} = 0.5$. Interestingly, the required values for $m_1$ and $m_2$ are nearly independent of the size of $\bD$.

\subsection{The proposed approach with real data}
The Extended Yale Face Database \cite{data1} consists of face images of 38 human subjects under different illumination. We select 50 images of a human subject. Fig. \ref{fig:face} shows few samples of the face images. It has been observed that these images roughly follow the low rank plus sparse model, and the low rank plus sparse matrix decomposition algorithms were successfully applied to such images to remove shadows and specularities \cite{lamport22}.
In addition, we add a sparse matrix with $\rho = 0.01$ to the face images.
We form a data matrix $\bD \in \mathbb{R}^{32256 \times 100}$ (32256 is the number of pixels per image) consisting of these 50 sparsely corrupted face images plus 50 randomly sampled images from the Caltech101 database \cite{data2} as outlying data points (50 images from random subjects). Fig. \ref{fig:random 101}  shows a subset of the images sampled from the Caltech101 database. The first 50 columns are the face images and the last 50 columns are the random images.
We have found that, for $ 1\leq i \leq 50 $, the average number of non-zero elements of $\bD \bz^{*}_i / \max ( | \bD \bz_i^{*}|)$ with absolute values greater than 0.1 is $0.05 N_2$ with standard deviation $0.04 N_2$. For $ 51\leq i \leq 100 $, the average number of non-zero elements of $\bD \bz^{*}_i /  \max ( | \bD \bz_i^{*}|)$ with absolute value greater than 0.1 is $0.37 N_2$ with standard deviation $0.14 N_2$. Thus, the non-face images can be identified with a proper threshold on the number of non-zero elements of $\bD \: \bz^{*}_i$. \\

\section{Appendix}
\label{sec:appendix}
In this section, we study a more general theoretical problem, dubbed null space learning problem, of which the sparse approximation problem is a special case. Similar to the model used in Section \ref{sec:SAtheory}, assume that matrix $\bA \in \mathbb{R}^{N_1 \times n}$ is a full rank matrix that can be expressed as $\bA = \bB + \bS$, where $\bS$ is a sparse matrix and the columns of $\bB$ are linearly dependent. If the CS of $\bB$ is not coherent with the standard basis, we expect the optimal point of
\begin{eqnarray}
\underset{{\bz}}{\min} \:\:
 \|  \bD \bz \|_0  \quad \text{subject to} \quad \| \bz \| = 1
 \label{eq:lzeronu00}
\end{eqnarray}
to lie in the null space of $\bB$ (i.e., $\bB \bz = \mathbf{0}$) because the CS of $\bB$ does not contain sparse vectors and the optimal point should cancel out the component corresponding to $\bB$. Accordingly, the optimal point of (\ref{eq:lzeronu00}) is equal to the optimal point of
\begin{eqnarray}
\underset{{\bz}}{\min} \:\:
 \|  \bS \bz \|_0  \quad \text{subject to} \quad \bB \bz = 0 \quad \text{and} \quad \| \bz \| = 1  \:.
 \label{eq:lzeronull0}
\end{eqnarray}
Therefore, the $\ell_0$-norm minimization problem can learn a direction in the null space of $\bB$. In fact, the optimization problem (\ref{eq:lzeronu00}) finds the most sparse vector in the CS\footnote{Interestingly, finding the most sparse vector in a linear subspace
has bearing on, and has been effectively used in, other
machine learning problems, including dictionary learning and
spectral estimation \cite{qu2014finding,spielman2012exact,demanet2013recovering}.} of $\bD$.

The optimization problem (\ref{eq:lzeronu00}) is non-convex. We relax the cost function using an $\ell_1$-norm and replace the quadratic constraint with a linear constraint as
\begin{eqnarray}
\underset{{\bz}}{\min} \:\:
 \|  \bD \bz \|_1  \quad \text{subject to}  \quad \bz^T \bv = 1 \:,
 \label{eq:analyze 1}
\end{eqnarray}
where $\bv \in \mathbb{R}^{n \time 1}$ is a fixed vector.
We refer to (\ref{eq:analyze 1}) as the convex null space learning optimization problem. If we set the constraint vector $\bv$ equal to $\be_i$, the null space learning optimization problem is equivalent to the sparse approximation optimization problem (\ref{eq:introduction_el1}), hence the latter is a special case of the former.
Note that in (\ref{eq:introduction_el1}), $\bz \in \mathbb{R}^{n-1}$ and in (\ref{eq:analyze 1}) $\bz \in \mathbb{R}^{n\times 1}$, yet the equivalence stems from the fact that if $\bz^{*}$ is the optimal point of (\ref{eq:introduction_el1}), then the optimal point of (\ref{eq:analyze 1}) with $\bv = \be_i$ will be equal to 
\[
[ \bz^{*}(1:i-1) \:\:, \:\: -1 \: \: , \:\: \bz^{*} (i:n-1) ]^T,
\]
where $\bz(i:j)$ denotes the elements of a vector $\bz$ from index $i$ to $j$. 

The following Theorem establishes sufficient conditions for the optimal point of the $\ell_1$-norm minimization problem (\ref{eq:analyze 1}) to lie in the null space of $\bB$. Before we state the theorem, let us define $\bz_o^{*}$ as the optimal point of
\begin{eqnarray}
\underset{{\bz}}{\min} \:\:
 \|  \bS \bz \|_1  \quad \text{subject to} \quad \bB \bz = 0 \quad \text{and} \quad \bz^T \bv = 1 \:.
 \label{eq:oracel 2}
\end{eqnarray}
 The sets $\calL_S$ and $\calL_{\bz_o^{*}}$ are defined similar to (\ref{eq:set_definition}).

\begin{theorem}
Suppose matrix $\bA \in \mathbb{R}^{N_1 \times n}$ is a full rank matrix that can be expressed as $\bA = \bB+\bS$. Define $\bz_o^{*}$ as the optimal point of (\ref{eq:oracel 2}), and define
\begin{eqnarray}
\alpha =  \sum_{ i }\sgn (\bs_{i}^T \bz_o^{*}) \: \ba_{i}  .
\end{eqnarray}
If
\begin{eqnarray}
\begin{aligned}
& \frac{1}{2} \underset{\delta \in \calR_b  \atop \| \delta \| = 1}{\inf} \left( \sum_{ i \in \calL_S }  | \bb_i^T \delta | - 2 \sum_{i \in \calL_S^c \cap {\cal L}_{\bz_o^{*}}}  \left| \delta^T \bb_i \right| \right) > \\
& \quad \quad \quad\quad\quad\quad\quad\quad  \sum_{i \in \calL_{\bz_o^{*}}}  \| \bs_{i} \| +  \| \alpha \| \: \:,\\
& \frac{\left\| \bv^T \bP_b \right\|}{2\| \bv^T\bR_b \|} \underset{\delta \in \calR_b  \atop \| \delta \|= 1}{\inf} \: \: \sum_{ i \in \calL_S }  | \bb_i^T \delta | > \\
& \quad \quad \quad\quad\quad\quad\quad\quad   \sum_{i \in \calL_{\bz_o^{*}}}  \left\| \bs_{i} \right\| + \| \alpha \|
\end{aligned}
\label{eq: lemma_inq}
\end{eqnarray}
 where $\bb_i$ and $\bs_i$ are the $i^{\text{th}}$ row of $\bB$ and $\bS$, respectively, and the subspace $\calR_b$ is the row space of $\bB$, then $\bz_o^{*}$ is the optimal point of (\ref{eq:analyze 1}).
\label{lm: two_ind}
\end{theorem}

The sufficient conditions (\ref{eq: lemma_inq}) reveal some interesting properties which merit some intuitive explanation provided next. 
According to Theorem \ref{lm: two_ind}, the following are important factors to ensure that the optimal point of (\ref{eq:analyze 1}) lies in the null space of $\bB$: \\
\textit{1. The CS of $\bB$ should not be coherent with the standard basis}:
Recall that we assume that $n \ll N_1$. Thus, if $\bS$ follows the Bernoulli model and $\rho$ and $n$ are sufficiently small, then $| \calL_S^c | \ll |\calL_S|$, in which case the LHS of (\ref{eq: lemma_inq}) will approximate the permeance statistic\cite{lerman2015robust} -- a measure of how well the rows of $\bB$ are distributed in the row space of $\bB$. The permeance statistic increases if the rows are more uniformly distributed in the row space.
But, if they are aligned along some specific directions, the permeance statistic tends to be smaller, wherefore the CS of $\bB$ will be more coherent with the standard basis.
This is in agreement with our initial intuition since linear combinations of the columns of $\bB$ are more likely to form sparse vectors when $\bB$ is highly coherent.
In other words, the coherence of $\bB$ would imply that $\bA \bz$ could be sparse even if $\bz$ does not lie in the null space of $\bB$.
\smallbreak
\noindent\textit{2. The matrix $\bS$ should be sufficiently sparse}: Per the first inequality of (\ref{eq: lemma_inq}), $\bS$ should be sufficiently sparse otherwise the cardinality of $\calL_S$ will not be sufficiently large. This requirement also confirms our initial intuition because the optimal point of (\ref{eq:analyze 1}) cannot yield a sparse linear combination even if it lies in the null space of $\bB$ unless $\bS$ is sparse. 
\smallbreak
%
\noindent\textit{3. The vector $\bv$ should not be too incoherent with $\calN_b$}: Recalling that $\bP_b$ and $\bR_b$ are orthonormal bases for the null space and row space of $\bB$, respectively,
the factor
\[
\frac{\left\| \bv^T \bP_b \right\|}{\| \bv^T\bR_b \|}
\]
on the LHS of the second inequality of (\ref{eq: lemma_inq}) unveils that the vector $\bv$  should be sufficiently coherent with $\calN_b$ (the null space of $\bB$) in order to ensure that the optimal point of (\ref{eq:analyze 1}) lies in $\calN_b$.
%
The intuition is that if $\bv$ has a very small projection on $\calN_b$ and the optimal point of (\ref{eq:analyze 1}) lies in $\calN_b$, then the optimal point of (\ref{eq:analyze 1}) should have a large Euclidean norm to satisfy the linear constraint of (\ref{eq:analyze 1}). In that sense, points lying in $\calN_b$ would be unlikely to attain the minimum of the objective function in (\ref{eq:analyze 1}). In addition, this coherency requirement implies that the optimal point of (\ref{eq:analyze 1}) is more likely to lie in $\calN_b$ when the rank of $\bB$ is smaller because the null space would have higher dimension, which makes $\bv$ more likely to be coherent with $\calN_b$.


\subsection{Proof of Lemma \ref{lm:yek}}
Lemma \ref{lm:yek} is a special case of Theorem \ref{lm: two_ind}. In order to prove Lemma \ref{lm:yek}, we make use of the following Lemmas from \cite{foucart2013mathematical,lerman2015robust} to lowerbound $\underset{\delta \in \calR_b  \atop \| \delta \|= 1}{\inf} \: \: \sum_{ i \in \calL_S }  | \bb_i^T \delta |$ and upperbound $\| \alpha \|$ in (\ref{eq: lemma_inq}).

\begin{lemma}
(Lower-bound on the permeance statistic from \cite{lerman2015robust}) Suppose that $\bg_1 , ... , \bg_n$ are i.i.d. random vectors uniformly distributed on the unit sphere $\mathbb{S}^{r - 1}$ in $\mathbb{R}^{r}$. When $r = 1$,
\begin{eqnarray}
\underset{\| \delta \| = 2}{\inf} \: \: \sum_{i = 1}^n \left| \mathbf{\delta}^T \bg_i \right| = 1.
\end{eqnarray}
When $r \ge 2$, for all $t \ge 0$,
\begin{eqnarray}
\underset{\| \delta \| = 2}{\inf} \: \: \sum_{i = 1}^n \left| \mathbf{\delta}^T \bg_i \right| > \sqrt{\frac{2}{\pi}} \frac{n}{\sqrt{r}} - 2\sqrt{n} -t \sqrt{\frac{n}{r -1 }}
\end{eqnarray}
 with probability at least $1 - exp(- t^2 / 2) \:$.
\end{lemma}

\begin{lemma}
If $\bg_1 , ... , \bg_n$ are i.i.d. random vectors uniformly distributed  on the unit sphere $\mathbb{S}^{r - 1}$ in $\mathbb{R}^{r}$, then
\begin{eqnarray}
\mathbb{P} \left( \left\| \sum_{i = 1}^n h_i \bg_i    \right\|  \ge \| \bh\| t   \right) \leq \exp \left( \frac{r}{2}(t^2 - log(t^2) -1) \right)
\end{eqnarray}
\end{lemma}
for all $t > 1$.

\subsection{Proof of Theorem \ref{lm: two_ind}}
We want to show that
\begin{eqnarray}
\underset{ \bv^T\bz = 1}{\arg\min} \: \:  \| \bA \: \bz \|_1 = \underset{\bz \in \calN_b \atop \bv^T \bz = 1}{\arg\min} \: \:  \| \bS \: \bz \|_1
\end{eqnarray}
Define $g(\delta) $ as
\begin{eqnarray}
g(\delta)  =  \| \bA \: (\bz_o^{*} - \mathbf{\delta})  \|_1 - \|  \bA \: \bz_o^{*} \|_1  \: .
\label{g_delta}
\end{eqnarray}
Since (\ref{eq:analyze 1}) is a convex optimization problem, it suffices to check that $g(\delta) \ge 0$ for every sufficiently small non-zero perturbation $\delta$ such that
\begin{eqnarray}
\delta^T \bv = 0 \: \: .
\label{condition_asli}
\end{eqnarray}
The conditions on $\delta$ is to ensure that $\bz_o^{*} - \mathbf{\delta}$ is a feasible point of (\ref{eq:analyze 1}).
If $\bz_o^{*}$ is the optimal point of (\ref{eq:oracel 2}), then the cost function of (\ref{eq:oracel 2}) is increased when we move from the optimal point along a feasible perturbation direction. Observe that $\bz_o^{*} - \delta_n$ is a
feasible point of (\ref{eq:oracel 2}), if and only if the perturbation $\delta_n$ satisfies
\begin{eqnarray}
\mathbf{\delta}_n^T \bv = 0 \: \: , \: \: \mathbf{\delta}_n \in \calN_b \:,
\label{condition_oracel}
\end{eqnarray}
where $\calN_b$ is the null space of $\bB$.
Therefore, for any non-zero $\mathbf{\delta}_n$ which satisfies (\ref{condition_oracel})
\begin{eqnarray}
\|\bS \: (\bz_o^{*} - \delta_n) \|_1 - \| \bS \: \bz_o^{*} \|_1 \ge 0 \: .
\label{strict_oracel}
\end{eqnarray}
When $\delta_n \to 0$, we can rewrite (\ref{strict_oracel}) as
\begin{align}
& \| \bS (\bz_o^{*} - \delta_n) \|_1 - \| \bS \bz_o^{*} \|_1 \nonumber\\
&= \sum_{i = 1}^{N_1} \left[ (\bs_i^T (\bz_o^{*} - \delta_n))^2 \right]^{1/2} -  \sum_{i = 1}^{N_1}  \left| \bs_i^T \bz_o^{*} \right| \nonumber \\
& =  \sum_{i = 1}^{N_1} \hspace{-.08cm}\left[ (\bs_i^T \bz_o^{*})^2 \hspace{-.08cm} - \hspace{-.08cm}  2 (\bs_i^T \bz_o^{*})(\mathbf{\delta}_n^T \bs_i) \hspace{-.1cm}+\hspace{-.1cm} (\mathbf{\delta}_n^T \bs_i)^2 \right]^{1/2} \hspace{-.15cm} - \hspace{-.1cm} \sum_{i = 1}^{N_1} \left| \bs_i^T \bz_o^{*} \right| \nonumber\\
&=  \sum_{ i \in \calL_{\bz_o^{*}}} \left| \delta_n^T \bs_i \right|
+ \sum_{i \in \calL_{\bz_o^{*}}^c} \left|\bs_i^T \bz_o^{*} \right| \bigg[ 1 - 2 \frac{\sgn (\bs_i^T \bz_o^{*})}{|\bs_i^T \bz_o^{*}|} (\mathbf{\delta}_n^T \bs_i) \nonumber\\
& \qquad\qquad\qquad\qquad +   \calO (\| \delta_n \|^2 )\bigg]^{1/2}  - \sum_{i \in \calL_{\bz_o^{*}}^c} \left| \bs_i^T \bz_o^{*} \right| \nonumber\\
&= \sum_{ i \in \calL_{\bz_o^{*}}} \left| \delta_n^T \bs_i \right| - \hspace{-0.25cm}\sum_{ i \in \calL_{\bz_o^{*}}^c }\sgn (\bs_i^T \bz_o^{*}) (\mathbf{\delta}_n^T \bs_i) + \mathcal{O} ( \|\mathbf{\delta_n}\|^2)
\end{align}
where the last identity follows from the Taylor expansion of the square root. Thus,
\begin{eqnarray}
\sum_{ i \in \calL_{\bz_o^{*}}} \left| \delta_n^T \bs_i \right| - \hspace{-0.25cm}\sum_{ i \in \calL_{\bz_o^{*}}^c }\sgn (\bs_i^T \bz_o^{*}) (\mathbf{\delta}_n^T \bs_i) + \mathcal{O} ( \|{\delta_n}\|^2)
\end{eqnarray}
has to be greater than zero for small $\delta_n$ which satisfies (\ref{condition_oracel}). Therefore,
\begin{eqnarray}
\begin{aligned}
&\sum_{ i \in \calL_{\bz_o^{*}}} \left| \delta_n^T \bs_i \right| - \hspace{-0.25cm}\sum_{ i \in \calL_{\bz_o^{*}}^c }\sgn (\bs_i^T \bz_o^{*}) (\mathbf{\delta}_n^T \bs_i) \ge 0 \: \: \: , \forall \: \delta_n \in \mathbb{R}^{M_1} \\
& \text{s.t.} \quad \delta_n^T \bv = 0 \: \:, \: \: \delta_n \in \calN_b \: .
\end{aligned}
\label{observation}
\end{eqnarray}

To simplify $g(\delta)$, we decompose $\delta$ into
\begin{eqnarray}
\delta = \delta_1 + \delta_2
\label{delta_decompose}
\end{eqnarray}
where $\delta_1 \in \calR_b$ and $\delta_2 \in \calN_b $, where $\calR_b$ is the row space of $\bB$. The vectors $\bz_o^{*}$ and $\delta_2$ lie in $\calN_b$. Therefore, we can expand $g(\delta)$ as
\begin{eqnarray}
\begin{aligned}
 & \| \bA (\bz_o^{*} - \mathbf{\delta})  \|_1 - \|  \bA \bz_o^{*} \|_1 = \|   \bS (\bz_o^{*} - \mathbf{\delta}) - \bB \delta_1 \|_1 - \|  \bS \bz_o^{*} \|_1 \\= 
 & \sum_{ i \in \calL_S } \big| \bb_i^T \delta_1 \big| + \sum_{i \in \: \calL_S^c \cap \calL_{\bz_o^{*}}} | \bs_i^T \delta +  \bb_i^T \delta | \\ 
 &+\sum_{ i \in \: \calL_S^c \cap \calL_{\bz_o^{*}}^c } \bigg| \bs_i^T (\bz_o^{*} - \mathbf{\delta}) - \bb_i^T \delta_1 \bigg|  - \|  \bS \bz_o^{*} \|_1.
\end{aligned}
\end{eqnarray}
Thus, it suffices to ensure that
\begin{eqnarray}
\begin{aligned}
&\sum_{ i \in \calL_S } \big| \bb_i^T \delta_1 \big| + \sum_{i \in \: \calL_S^c \cap \calL_{\bz_o^{*}}} | \bs_i^T \delta   | - \sum_{i \in \: \calL_S^c \cap \calL_{\bz_o^{*}}} |  \bb_i^T \delta |  \\
 &+\sum_{ i \in \: \calL_S^c \cap \calL_{\bz_o^{*}}^c } \bigg| \bs_i^T (\bz_o^{*} - \mathbf{\delta}) - \bb_i^T \delta_1 \bigg|  - \|  \bS \bz_o^{*} \|_1
\end{aligned}
\label{upper_1}
\end{eqnarray}
is non-negative since (\ref{upper_1}) is a lower bound on $g (\delta)$.
In addition, as $\delta \to 0$,
\begin{eqnarray}
\begin{aligned}
 &\sum_{ i \in \: \calL_S^c \cap \calL_{\bz_o^{*}}^c } \bigg| \bs_i^T (\bz_o^{*} - \mathbf{\delta}) - \bb_i^T \delta_1 \bigg|  - \|  \bS \bz_o^{*} \|_1  \\
&  = - \hspace{-0.25cm}\sum_{i \in \: \calL_S^c \cap \calL_{\bz_o^{*}}^c }\sgn (\bs_i^T \bz_o^{*}) (\mathbf{\delta}^T \ba_i) + \mathcal{O} (\| \delta^2 \|) \: .
\end{aligned}
\label{simplify1}
\end{eqnarray}
Therefore, according to (\ref{g_delta}), (\ref{simplify1}) and (\ref{upper_1}), it is enough to show that
\begin{eqnarray}
\begin{aligned}
& \sum_{ i \in \calL_S } \big| \bb_i^T \delta_1 \big|  -  \sum_{i \in \: \calL_S^c \cap \calL_{\bz_o^{*}}} |  \bb_i^T \delta | \\
& + \sum_{ i \in \calL_{\bz_o^{*}}} \left| \delta^T \bs_i \right| - \hspace{-0.1cm}\sum_{}\sgn (\bs_i^T \bz_o^{*}) (\mathbf{\delta}^T \ba_i) \ge 0 \:,
\end{aligned}
\end{eqnarray}
for every $\delta \neq 0$ which satisfies (\ref{condition_asli}). Define
\begin{eqnarray}
\alpha =  \sum\sgn (\bs_i^T \bz_o^{*})  \ba_i.
\label{alpha}
\end{eqnarray}
Therefore, to show that $g(\delta)$ is non-negative, it suffices to ensure that
\begin{eqnarray}
\begin{aligned}
& \frac{1}{2} \sum_{ i \in \calL_S } \big| \bb_i^T \delta_1 \big|  - \sum_{ i \in \: \calL_S^c \cap \calL_{\bz_o^{*}}} | \bb_i^T \delta_1 | - \sum_{i \in \calL_{\bz_o^{*}}}  \left| \delta_1^T \bs_i \right| > \delta_1^T \alpha \\
& \frac{1}{2} \sum_{ i \in \calL_S } \big| \bb_i^T \delta_1 \big|  > \delta_2^T  \alpha - \sum_{i \in \calL_{\bz_o^{*}}}  \left| \delta_2^T \bs_i \right|
\end{aligned}
\label{cond_lemma}
\end{eqnarray}
For the first inequality of (\ref{cond_lemma}), it is enough to ensure that a lower bound on the LHS is greater than an upper bound on the RHS. Thus, it suffices to have
\begin{eqnarray}
\begin{aligned}
& \frac{1}{2} \underset{\delta \in \calR_b  \atop \| \delta \| = 1}{\inf} \left( \sum_{ i \in \calL_S }  | \bb_i^T \delta | - 2 \sum_{i \in \: \calL_S^c \cap \calL_{\bz_o^{*}}}  \left| \delta^T \bb_i \right| \right) \\
& > \underset{\delta \in \calR_b \atop \| \delta \| = 1}{\sup} \sum_{i \in \calL_{\bz_o^{*}}}  \left| \delta^T \bs_i \right| +    \underset{\delta \in \calR_b  \atop \| \delta \| = 1}{\sup} \delta^T \alpha.
\end{aligned}
\label{sup_min1}
\end{eqnarray}
Thus, the first inequality of (\ref{eq: lemma_inq}) guarantees the first inequality of (\ref{cond_lemma}).

Observe that the second inequality of (\ref{cond_lemma}) is homogeneous in $\delta$ since
\begin{eqnarray}
\delta_1^T \bv = - \delta_2^T \bv \:.
\end{eqnarray}
We scale $\delta$ such that $\delta_1^T \bv = - \delta_2^T \bv = 1$. To ensure that the second inequality of (\ref{cond_lemma}) is satisfied, it is enough to show that
\begin{eqnarray}
\begin{aligned}
&\frac{1}{2} \underset{\delta_1 \in \calR_b  \atop \delta_1^T \bv = 1}{\inf} \: \:  \sum_{ i \in \calL_S }  | \bb_i^T \delta_1 | \\
& > \underset{\delta_2 \in \calN_b \atop \delta_2^T \bv = 1}{\sup} \: \: \left( \delta_2^T  \alpha -  \sum_{i \in \calL_{\bz_o^{*}}}  \left| \delta_2^T \bs_i \right|  \right).
\label{linear_cond}
\end{aligned}
\end{eqnarray}
Let us decompose $\delta_2$ into
\begin{eqnarray}
\delta_2 = \delta_{2p} + \delta_{2v}
\end{eqnarray}
where
\begin{eqnarray}
\delta_{2p} = (\bI - \bv^{'} (\bv^{'})^T) \delta_2 \: \: , \: \: \delta_{2v} = \bv^{'} (\bv^{'})^T \delta_2
\end{eqnarray}
and $\bv^{'}$ is defined as
\begin{eqnarray}
\bv^{'} = \bP_b \bP_b^T \bv  \:/ \: \| \bP_b \bP_b^T \bv \|,
\end{eqnarray}
where $\bP_b$ was defined as an orthonormal basis for $\calN_b$. For the second inequality, it is enough to show that the LHS of (\ref{linear_cond}) is greater than
\begin{eqnarray}
\underset{\delta_2 \in \calN_b  \atop \delta_2^T \bv = 1}{\sup}  \left( \delta_{2p}^T  \alpha + \delta_{2v}^T  \alpha - \sum_{i \in \calL_{\bz_o^{*}}}  \left| \delta_{2p}^T \bs_i \right|   +  \sum_{i \in \calL_{\bz_o^{*}}}  \left| \delta_{2v}^T \bs_i \right|  \right).
\end{eqnarray}
According to the definition of $\bv^{'}$ and $\delta_{2p}$,
\begin{eqnarray}
\delta_{2p} \in  \calN_b \quad \text{and} \quad \delta_{2p}^T \bv = 0 \: .
\end{eqnarray}
Therefore, according to (\ref{observation}), $$\delta_{2p}^T \alpha - \sum_{i \in \calL_{\bz_o^{*}}}  \left| \delta_{2p}^T \bs_i \right| \leq 0 \: .$$
Thus, it is enough to show that the LHS of (\ref{linear_cond}) is greater than
\begin{eqnarray}
\begin{aligned}
& \underset{\delta_2 \in \calN_b  \atop \delta_2^T \bv = 1}{\sup}  \: \: \left( \delta_{2v}^T  \alpha +   \sum_{i \in \calL_{\bz_o^{*}}}  \left| \delta_{2v}^T \bs_i \right| \right) \\
& = \frac{1}{\left\| \bv^T \bP_b \right\|} \left( \left| \alpha^T \bv^{'} \right| + \sum_{i \in \calL_{\bz_o^{*}}}  \left| \bs_i^T \bv^{'} \right| \right)   \:.
\end{aligned}
\label{dovom_1}
\end{eqnarray}
In addition, the LHS of (\ref{linear_cond}) can be simplified as
\begin{eqnarray}
\begin{aligned}
&\frac{1}{2} \underset{\delta_1 \in \calR_b  \atop \delta_1^T \bv = 1}{\inf} \: \:  \sum_{ i \in \calL_S }  | \bb_i^T \delta_1 | \ge \\
 &\frac{1}{2\| \bv^T\bR_b \|} \underset{\delta_1 \in \calR_b  \atop \| \delta_1 \|= 1}{\inf} \: \: \sum_{ i \in \calL_S }  | \bb_i^T \delta_1 | \:.
\end{aligned}
\label{dovom_2}
\end{eqnarray}
According to (\ref{linear_cond}), (\ref{dovom_1}) and (\ref{dovom_2}), the second inequality of (\ref{eq: lemma_inq}) guarantees that the second inequality of (\ref{cond_lemma}) is satisfied.


\bibliographystyle{IEEEtran}
\bibliography{IEEEabrv,bibfile}


\end{document}